\pdfoutput=1

\documentclass[11pt]{article}

\usepackage[]{naacl2021}

\usepackage{times}
\usepackage{latexsym}

\usepackage[]{xcolor}
\definecolor{OliveGreen}{RGB}{34,139,34}

\usepackage{graphicx}

\usepackage{pifont}

\usepackage[T1]{fontenc}

\usepackage[utf8]{inputenc}
\usepackage{todonotes}

\usepackage{microtype}

\title{A dissemination workshop for introducing young Italian students to NLP}

\author{
  Lucio Messina \\
  Independent Researcher\\
 \texttt{lucio.messina@autistici.org} \\\And
  Lucia Busso \\
  Aston University \\
 \texttt{l.busso@aston.ac.uk} \\\AND
  Claudia Roberta Combei \\
  University of Bologna\\
 \texttt{claudiaroberta.combei@unibo.it~~~} \And
  Ludovica Pannitto \\
  University of Trento \\
  \texttt{ludovica.pannitto@unitn.it} \\
 \AND
 Alessio Miaschi \\
  University of Pisa\\
 \texttt{alessio.miaschi@phd.unipi.it~~~} \\\And
 Gabriele Sarti \\
 ~~~University of Trieste\\
 \texttt{~~~gsarti@sissa.it} \\\And
 Malvina Nissim \\
  University of Groningen\\
 \texttt{m.nissim@rug.nl} \\
 }

\begin{document}
\maketitle
\begin{abstract}

We describe and make available the game-based material developed for a laboratory run at several Italian science festivals to popularize NLP among young students. 

\end{abstract}

\section{Introduction}

The present paper aims at describing in detail the teaching materials developed and used for a series of interactive dissemination workshops on NLP and computational linguistics\footnote{In this discussion, and throughout the paper, we conflate the terms Natural Language Processing and Computational Linguistics and use them interchangeably.}. These workshops were designed and delivered by the authors on behalf of the Italian Association for Computational Linguistics (AILC, \url{www.ai-lc.it}), with the aim of  popularizing Natural Language Processing (NLP) among young Italian students (13+) and the general public. The workshops were run in the context of nationwide popular science festivals and open-day events both onsite (at \textit{BergamoScienza}, and Scuola Internazionale Superiore di Studi Avanzati [SISSA], Trieste) and online (at \textit{Festival della Scienza di Genova}, the BRIGHT European Researchers' Night, high school ITS Tullio Buzzi in Prato and the second edition of \textit{Science Web Festival}, engaging over 700 participants in Center and Northern Italy from 2019 to 2021.\footnote{Links to events are in the repository's README file.} 

The core approach of the workshop remained the same throughout all the events. However, the materials and activities were adapted to a variety of different formats and time-slots, ranging from 30 to 90 minutes. 
We find that this workshop -- thanks to its modular nature -- can fit different target audiences and different time slots, depending on the level of interactive engagement required from participants and on the level of granularity of the presentation itself.
Other than on the level of engagement  expected of participants, time required can also vary depending on the participants' background and metalinguistic awareness. 

Our interactive workshops took the form of modular games where participants, guided by trained tutors, acted as if they were computers that had to recognize speech and text, as well as to generate written sentences in a mysterious language they knew nothing about.

The present contribution only describes the teaching materials and provide a general outline of the activities composing the workshop. For a detailed discussion and reflection on the workshop genesis and goals and on how it was received by the participants see \cite{pannitto2021teaching}.

The teaching support consist in an interactive presentation plus hands-on material, either in hard-copy or digital form.
We share a sample presentation\footnote{\url{https://docs.google.com/presentation/d/1ebES_K8o3I2ND_1iMyBlQmrH733eQ6m_K8a0tON8reo/edit?usp=sharing}} and an open-access repository\footnote{\url{https://bitbucket.org/melfnt/malvisindi}} containing both printable materials to download and scripts to reproduce them on different input data.

\section{Workshop and materials}

The activity contains both more theoretical and hands-on parts, which are cast as games.

\paragraph{Awareness}
The first part consists of a brief introduction to (computational) linguistics, focusing on some common misconceptions (slides 3-5), and on examples of linguistic questions (slides 6-12). Due to their increasing popularity, we chose vocal assistants as practical examples of NLP technologies accounting for how humans and machines differ in processing speech in particular and language in general (slides 20-39).

\paragraph{Games}

The core of the activity is inspired by the \textit{word salad} puzzle \citep{radev2013puzzles} and is organized as a game revolving around a fundamental problem in NLP: given a set of words, participants are asked to determine the most likely ordering for a sentence containing those words.  This is a trivial problem when approached in a known language (i.e., consider reordering the tokens \textit{garden, my, is, the, in, dog}), but an apparently impossible task when semantics is not accessible, which is the most common situation for simple NLP algorithms.

\begin{figure}[!ht]
    \centering
    \includegraphics[width=0.45\textwidth]{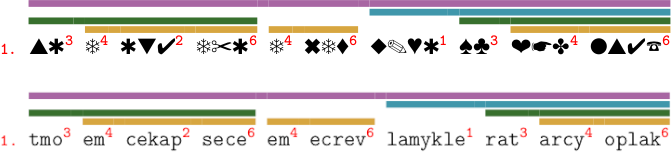}
	\caption{The first sentence of the corpus \texttt{"on a snowy day a queen was sewing by her window"} translated using \textit{DING}s (above) and using non-words (below)}
	\label{fig:transliterated_corpus}
\end{figure}

To make participants deal with language as a computer would do, we asked them to compose sentences using tokens obtained by transliterating and annotating 60 sentences from the well known fairy tale "Snow White" to a set of symbols. 
We produced two possible versions of the masked materials: either replacing each word with a random sequence of \textit{DING}s (e.g. \ding{100}\ding{169}\ding{220}\ding{156}\ding{61} for the word \textit{morning}) or replacing them with a corresponding non-word (for example \texttt{croto} for the word \textit{morning}). 
The grammatical structure of each sentence is represented by horizontal lines on top of it representing phrases (such as noun or verb phrases), while the parts of speech are indicated by numbers from 0 to 9 placed as superscripts on each word (Figure~\ref{fig:transliterated_corpus}).

Participants were divided into two teams, one team would be working on masked Italian and the other on masked English. Both teams were given the corpus in A3 format and were told that the texts are written in a fictional language.

\begin{figure}
    \centering
    \includegraphics[width=0.4\textwidth]{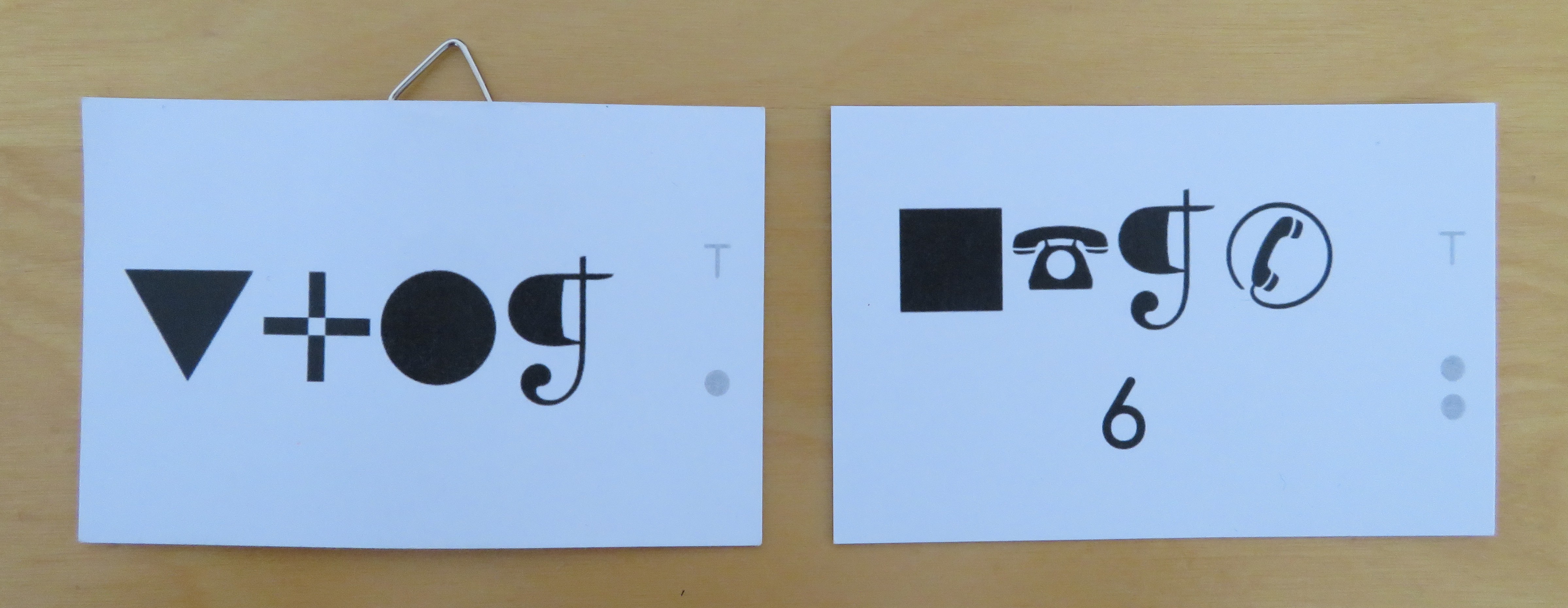}
    \caption{Example cards, both showing a word. Left: a card for the first activity, with a button loop to thread it in a sentence. Right: a card for the second activity, with part of speech (number) at the bottom.}
    \label{fig:tokens}
\end{figure}

Two activities were then run, focusing on two different algorithms for sentence generation.
In the first, participants received a deck of cards each equipped with a button~loop (Figure~\ref{fig:tokens}) and showing a token from the corpus. Participants had to create new valid sentences by rearranging the cards according to the bigrams' distribution in the corpus.
Using the \textit{bracelet method} (slides 52-61), they could physically thread tokens into sentences.

\begin{figure}
    \centering
    \includegraphics[width=0.3\textwidth]{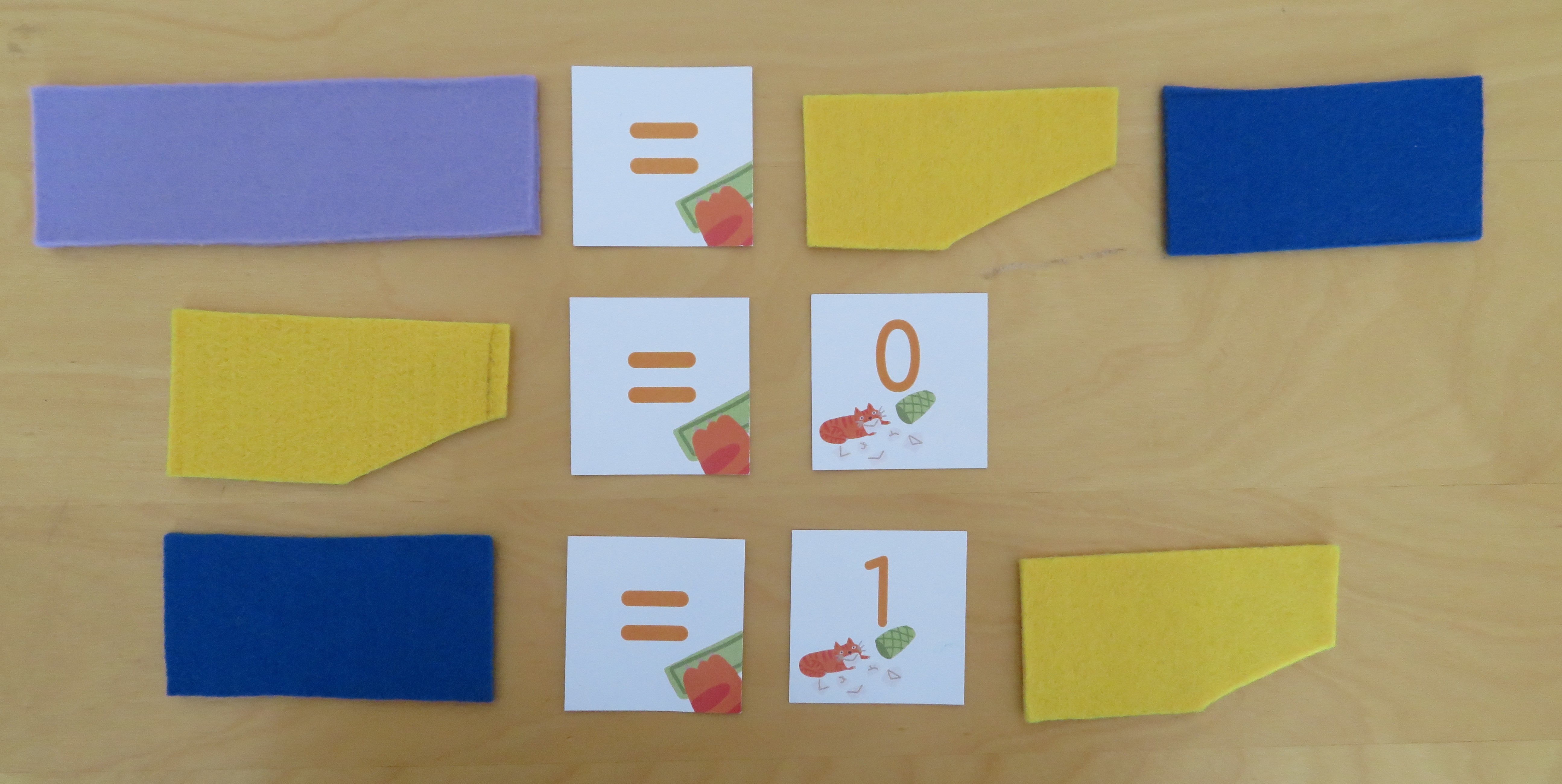}
    \caption{Possible rules extracted from the corpus. Each rule is made of felt strips for phrases, cards with numbers for parts of speech, and ``$=$'' cards.}
    \label{fig:rules}
\end{figure}

In the second activity (slides 63-92), the participants extracted grammatical rules from the corpus, and used them to generate new sentences.
In order to write the grammar, participants were given felt strips reproducing the colors of the annotation, a deck of cards with numbers (identifying parts of speech) and a deck of ``$=$'' symbols (Figure~\ref{fig:rules}).
With a new deck of words (Figure~\ref{fig:tokens}), not all present in the corpus, participants had to generate a sentence using the previously composed rules.

\paragraph{Reflection and Outlook}

\begin{figure}
    \centering
    \includegraphics[width=0.3\textwidth]{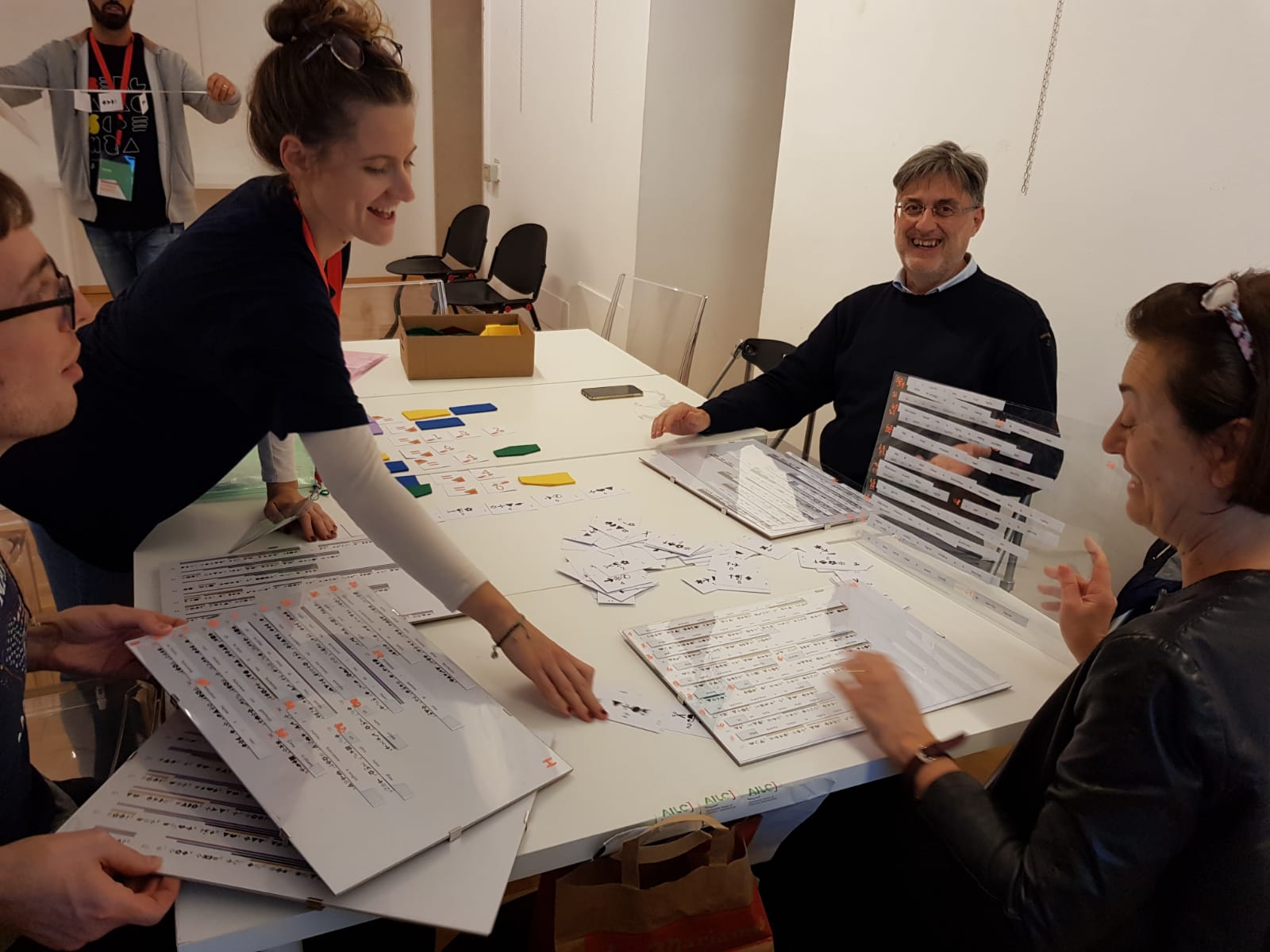}
    \caption{A session of the workshop: the original language of the corpus has just been revealed by superimposing plexiglass supports on the corpus tables.}
    \label{fig:plexiglass}
\end{figure}

By superimposing a plexiglass frame on the A3 corpus pages (Figure~\ref{fig:plexiglass}), the true nature of the corpora was eventually revealed. The participants could see the original texts (in Italian and English) and translate the sentences they had created previously. 

The activity ended with a discussion of  recent NLP technologies and their commercial applications (slides~93-96), and of what it takes to become a computational linguist today (slides~97-99).

\section{Activity Preparation}

The preparation of the activity consists of several steps:
(1) creating and tagging the corpora with morpho-syntactic and syntactic categories as described in the repository;
(2) choosing the words to include in the card decks: these must be manually selected but scripts are provided to generate  possible sentences based on bigram co-occurrences, and to extract all the possible grammar rules present in the annotation;
(3) when the produced sentences and grammar are satisfactory, scripts are provided to generate (i) the printable formats of corpora and decks of cards, (ii) a dictionary to support the translation of sentences in the last part of the workshop, and (iii) clear-text corpora;
(4) sentences from the clear-text corpora have to be manually cut and glued on a transparent support that can be superimposed on the printed corpora to {reveal} the sentences;
(5) finally, some manual work is necessary: producing strips of felt or any material with the same colors used in the corpus; cutting threads; attaching a button~loop to the relevant cards, etc.

\section{Reusability}

In the spirit of open science and to encourage the popularization of NLP, the teaching materials and source code are freely available in our repository (see footnotes~2-3 for the links). The print-ready material is released under CC BY-NC;  the source code is distributed under the GNU gpl license version~3.
All scripts work for Python versions 3.6 or above and the overall process requires \texttt{python3, lualatex, pdftk} and \texttt{pdfnup} as detailed in the \texttt{README.md} file in the repository, which contains all necessary instructions.

\section*{Acknowledgements}
We would like to thank the board of the Italian Association for Computational Linguistics (AILC) for the support given to the workshop development and delivery. We are also grateful to BergamoScienza, Festival della Scienza di Genova, Science Web Festival for hosting the activity during the festivals; to ILC-CNR ``Antonio Zampolli" and ColingLab (University of Pisa) for hosting us during the European Night of Research; and to the Scuola Internazionale Superiore di Studi Avanzati (SISSA) and ITI Tullio Buzzi for hosting our activities with their students. 
We also thank Dr. Mirko Lai, who has collaborated on the development of the web interface for the online versions of our activity.

\bibliography{anthology,custom}
\bibliographystyle{acl_natbib}

\end{document}